\documentclass[letterpaper]{article} 
\usepackage{aaai2026}  
\usepackage{times}  
\usepackage{helvet}  
\usepackage{courier}  
\usepackage[hyphens]{url}  
\usepackage{graphicx} 
\usepackage{natbib}  
\usepackage{caption} 
\usepackage{xcolor} 
\usepackage{algorithm}
\usepackage{xspace}
\usepackage{algorithmic}
\usepackage{adjustbox}
\usepackage{pifont} 
\usepackage{booktabs}
\usepackage{amssymb} 
\usepackage{amsmath}

\usepackage{amsthm}

\usepackage{color, colortbl}
\usepackage{bbding}
\usepackage{pifont}
\usepackage{amssymb}
 
\newcommand{\xmarkg}{\textcolor{gray}{\ding{55}}\xspace}%
\newcommand{\cmark}{\ding{51}}

\usepackage{newfloat}
\usepackage{listings}
\DeclareCaptionStyle{ruled}{labelfont=normalfont,labelsep=colon,strut=off} 
\floatstyle{ruled}
\newfloat{listing}{tb}{lst}{}
\floatname{listing}{Listing}

\title{Lifelong Language-Conditioned Robotic Manipulation Learning}
\author{
    Xudong Wang\textsuperscript{\rm 1,2}\equalcontrib, Zebin Han\textsuperscript{\rm 3,1,4}\equalcontrib, Zhiyu Liu\textsuperscript{\rm 1,2}, Gan Li\textsuperscript{\rm 3,1}, Jiahua Dong\textsuperscript{\rm 5}, Baichen Liu\textsuperscript{\rm 1}, \\ Lianqing Liu\textsuperscript{\rm 1}, Zhi Han\textsuperscript{\rm 1}\thanks{Corresponding Author}
}
\affiliations{
    \textsuperscript{\rm 1}State Key Laboratory of Robotics and Intelligent Systems, Shenyang Institute of Automation, Chinese Academy of Sciences\\
    \textsuperscript{\rm 2}University of Chinese Academy of Sciences\\
    \textsuperscript{\rm 3}North University of China\\
    \textsuperscript{\rm 4}Southeast University\\
    \textsuperscript{\rm 5}Mohamed bin Zayed University of Artificial Intelligence\\
     \{wangxudong,liuzhiyu,liubaichen,lqliu,hanzhi\}@sia.cn.com, \{zebinhan04,dongjiahua1995,gannonligan\}@gmail.com \
}

\usepackage{bibentry}

\begin{document}

\maketitle

\begin{abstract}
Traditional language-conditioned manipulation agent sequential adaptation to new manipulation skills leads to catastrophic forgetting of old skills, limiting dynamic scene practical deployment. In this paper, we propose \textit{SkillsCrafter}, a novel robotic manipulation framework designed to continually learn multiple skills while reducing catastrophic forgetting of old skills. Specifically, we propose a \textit{Manipulation Skills Adaptation} to retain the old skills knowledge while inheriting the shared knowledge between new and old skills to facilitate learning of new skills. Meanwhile, we perform the singular value decomposition on the diverse skill instructions to obtain common skill semantic subspace projection matrices, thereby recording the essential semantic space of skills. To achieve forget-less and generalization manipulation, we propose a \textit{Skills Specialization Aggregation} to compute inter-skills similarity in skill semantic subspaces, achieving aggregation of the previously learned skill knowledge for any new or unknown skill. Extensive experiments demonstrate the effectiveness and superiority of our proposed SkillsCrafter. 

\end{abstract}

\begin{links}
\link{Code}{https://skillscrafter-lifelong.github.io/}
\end{links}

\section{Introduction}

Robotic manipulation has seen significant advancements facilitated by large vision-language models, enabling robots to interpret and execute complex instructions grounded in visual perception \cite{FAAI, Tmech, MIR}. In particular, language-conditioned robotic manipulation (LCRM) has emerged as a powerful paradigm that allows robots to follow open-ended natural language instructions. Recent manipulation works such as LH-VLA~\cite{R6}, SRT-H~\cite{SRT-H}, and PaLM-E~\cite{palme} show impressive performance on LCRM in simulated and real-world environments.

\begin{figure*}[!t]
\centering
\includegraphics[width=1\linewidth]{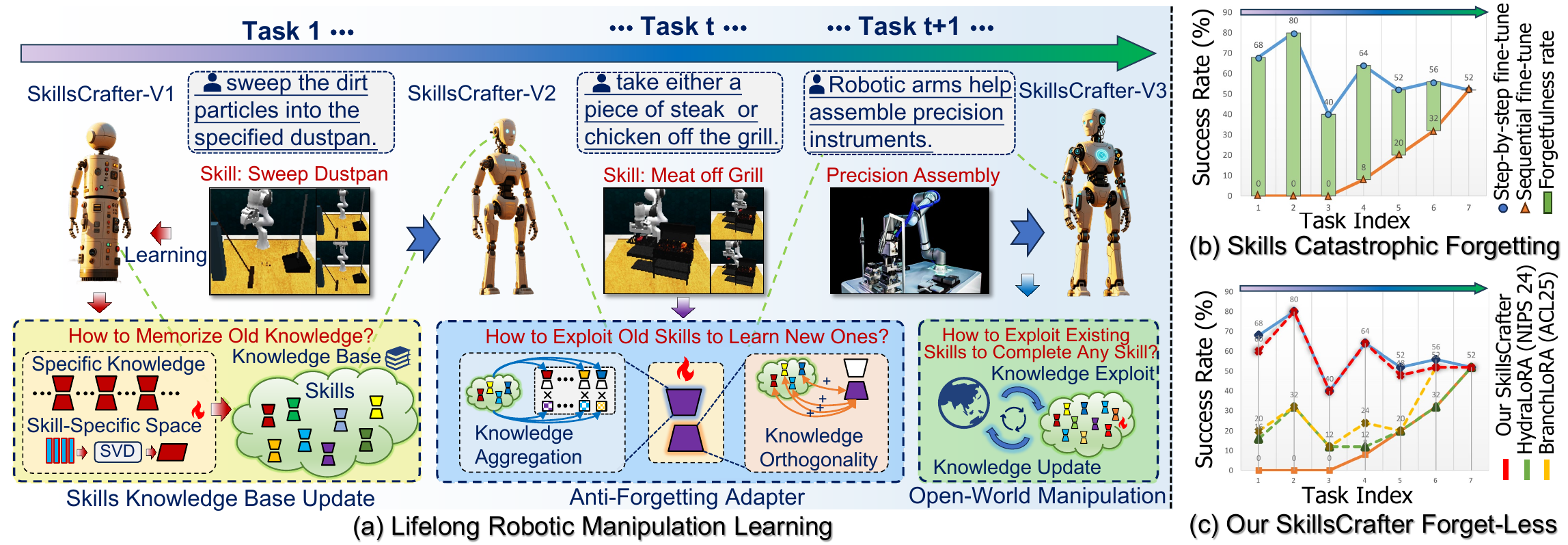}
\caption{The proposed Lifelong  Robotic Manipulation task. (a) SkillsCrafter is able to evolve and learn new skills based on learned skills. It maintains a skill knowledge base that stores learned knowledge to allow for efficient learning of new skills using shared knowledge of old skills, and for performing any unknown skill in the open world using previously learned knowledge. (b) Skills catastrophic forgetting under lifelong learning settings. (c) Our SkillsCrafter has a better anti-forgetting performance.
}
\label{fig_1}
\vspace{-2mm}
\end{figure*}

Despite these advances, enabling robots to continually acquire new skills without forgetting previously learned skills remains a major challenge \cite{ICEPE}. Most existing manipulation agents require task-specific fine-tuning, which leads to catastrophic forgetting of old skills when adapting to new skills~\cite{R46}. This severely hinders the practical deployment of manipulation robots in dynamic environments where new skills are integrated over time without compromising existing competencies. As shown in Figure~\ref{fig_1} (b), naively sequential fine-tuning an agent on skills results in a significant catastrophic forgetting on learned skills.

To address this problem, a straightforward method is to store separate models with full parameter checkpoints for each skill, which incurs substantial memory and computational overhead, making them impractical for long-term deployment. Recent large models leverage LoRA~\cite{lora}, a parameter-efficient tuning technique that injects small trainable low-rank matrices into a frozen base model. This fine-tuning method reduces storage by only maintaining lightweight LoRA adapters per skill task. However, it often treats each skill independently, ignoring the skill-shared knowledge (\emph{e.g.}, basic actions like grasping, rotating, and pressing) across related manipulation skills. Moreover, selecting and loading the corresponding adapter at inference time typically requires manual intervention, which limits scalability. MoE-LoRA~\cite{moelora}, HydraLoRA~\cite{HydraLoRA}, Branch-LoRA~\cite{BranchLoRA}, \emph{et al.}, introduce an automatic routing between multiple LoRA experts via mixture-of-experts gating. However, these methods require pre-defining the number of experts, which risks overfitting or under-utilization when the number is suboptimal. These limitation highlights a critical need for more effective methods to continue robotic manipulation skills acquisition. 

To handle the above practical scenarios, we introduce a new practical task named \textit{\underline{L}ifelong \underline{L}anguage-\underline{C}onditioned \underline{R}obotic \underline{M}anipulation (LLCRM)}. In this LLCRM setting, as shown in Figure~\ref{fig_1} (a), a manipulation robotic agent continuously upgrades and evolves based on lifelong learning with multi-skills. LLCRM aims to enable agents to accumulate, reuse, and refine manipulation skills over time. Inspired by \textit{how humans leverage analogies and prior experiences when acquiring new skills, we argue that efficient lifelong manipulation learning requires both skills \textbf{inheritance} and \textbf{integration}.} Thus we identify two core challenges in LLCRM:
\begin{enumerate}
    \item How to explore and exploit the shared-knowledge and specific-knowledge between manipulation skills?
    \item How to adaptively and flexibly aggregate the acquired knowledge for dealing with new or unknown skills?
\end{enumerate}

To resolve the challenges in the LLCRM task, we propose SkillsCrafter, a novel lifelong manipulation framework that explicitly reuses inter-skill knowledge. SkillsCrafter maintains a skill knowledge base that stores learned skill knowledge to allow for efficient learning of new skills using previously learned skill knowledge, even for performing unknown skills in the open world. It comprises two core components: i). Manipulation Skills Adaptation (MSkA), which continually learns new skills while preserving old ones, by separating LoRA into skill-shared and skill-specific components with skills-shared knowledge inheritance and orthogonal constraints; ii). Skills Specialization Aggregation (SkSA), which computes inter-skills similarity in skill semantic subspace to achieve aggregation of the previously learned skills knowledge. And we also analyze the efficiency of our skills-shared knowledge inheritance strategy from theory. Extensive experiments in both simulated and real-world deployments have demonstrated that SkillsCrafter enables robots to learn new skills without forgetting previous ones, achieving the efficient SOTA performance in lifelong manipulation. The main contributions are outlined below:
\begin{itemize}
    \item We introduce Lifelong Language-Conditioned Robotic Manipulation, a new task that enables agents to evolve and learn new skills based on learned skills. We also provide a new lifelong robotic manipulation benchmark for training and evaluating the lifelong manipulation task.

    \item To achieve efficient skills lifelong learning, we propose a Manipulation Skills Adaptation to retain old skills knowledge while inheriting the shared knowledge between new and old skills, to facilitate new skills learning.

    \item For anti-forgetting manipulation, we propose a Skills Subspaces Specialization Aggregation to compute inter-similarity in skill semantic subspaces, to achieve skills knowledge aggregation for manipulation skill inference. 
    
\end{itemize}

\section{Problem Formulation}

\noindent \textbf{Preliminary:} In language-conditioned robotic manipulation task \cite{LCRM1, LCRM2}, an embodied agent is required to understand user language instruction $\mathcal{I}$ to complete a series of manipulation skills. Following LLARVA \cite{llarva}, we use a pre-trained LLM, i.e., LLaMA 2 \cite{llama2} with a tokenizer $e$, as the basic robotic agent $\mathcal{F}$. And to process the vision observation $\mathcal{O}$, we encode it using the CLIP vision-encoder $\mathcal{V}(\mathcal{O})$. The agent model architecture is similar to the multimodal large language model LLaVA \cite{llava}. At each time $i$, the agent reasons the instruction $\mathcal{I}^{i}$ combined with the observation $\mathcal{O}^{i}$ to generate the next action as follows: $R^{i}=\mathcal{F}(\mathcal{V}(\mathcal{O}^{i}), e(\mathcal{I}^{i}))$. However, as shown in figure \ref{fig_1} (b), the agent's adaptation to a new skill $S_{t}$ causes catastrophic forgetting of the old skills $\{S_{1}, S_{2}, ..., S_{t-1}\}$, which limits its flexible deployment, and the existing robotic manipulation agents struggle to grow in continual  multi-skills. 

\begin{figure*}[!t]
\centering
\includegraphics[width=0.95\linewidth]{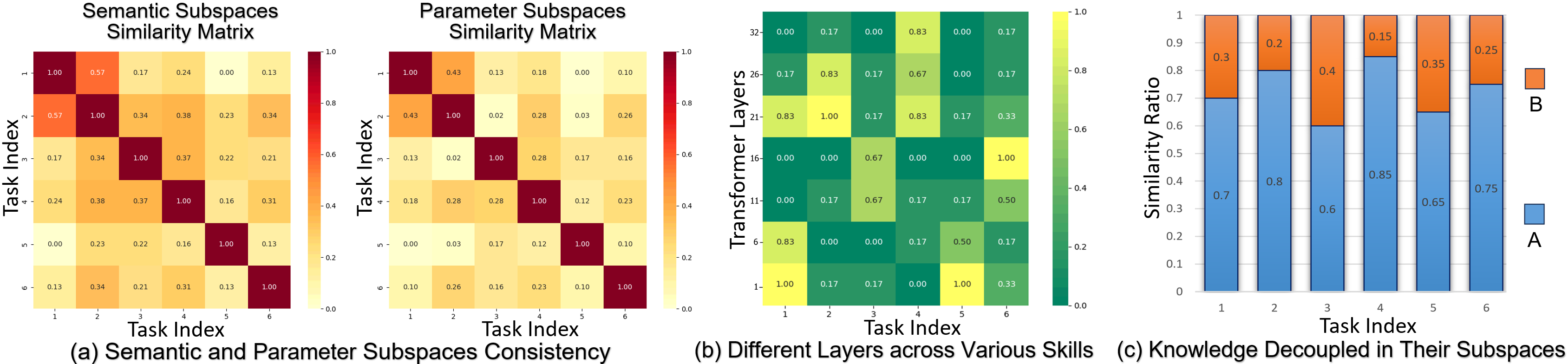}
\caption{Illustration of the three sets of observation. (a) Subspaces Consistency: The semantic space can be used to associate with the parameter subspace. (b) Different Layers across Various Skills: The importance of different layers is not the same for different skills. (c) Knowledge Decoupled: $\textbf{A}$ tends to learn shared knowledge, while $\textbf{B}$ to learn specific knowledge, naturally.
}
\label{fig_AF1}
\vspace{-2mm}
\end{figure*}

\noindent \textbf{Problem Definition:} To address the above lifelong learning challenges, we introduce a new problem setting, Lifelong Language-Conditioned Robotic Manipulation (LLCRM). We define a multi-skills sequence set $\mathcal{S}=\{S_{1}, S_{2}, ..., S_{t}\}$, where $t$-th skill $S_{t}=\{\mathcal{O}_{t},\mathcal{I}_{t}\}$ comprises vision observation $\mathcal{O}_{t} = \{o_{1},o_{2},...,o_{N}\}$, and an instruction $\mathcal{I}_{t}$. Agent $\mathcal{F}$ is required to learn all the skills $\mathcal{S}$ sequentially, and all skills are tested after learning is complete. And in LLCRM settings, the skill-id $t$ is agnostic during the testing phase. $\mathcal{I}_{t}$ of $S_{t}$ does not overlap with any previous skills: $\mathcal{I}_{t} \bigcap (\bigcup^{t-1}_{j=1}\mathcal{I}_{j})\!=\!\emptyset$. The LLCRM task aims to continually learn a sequence of new skills while alleviating the forgetting of old skills. In addition, to better explore and exploit skill knowledge, we define the \textit{Skill Parameter Subspace} as the skill parameter learned by the agent to adapt to diverse new skills, the \textit{Skill Semantic Subspace} as the semantic latent space embedding of user instructions describing skills.

\section{Preliminary Study}

In this section, we present our key observations regarding the two core challenges outlined in the Introduction. Various robotic skills not only have their specific knowledge, but also have shared knowledge. For example, ``unscrewing a bottle cap" and ``turning on a tap" both have rotating actions, and even ``catching a bottle" and ``roasting a chicken", which seem to be different skills, have commonalities in grasping actions. We use LoRA \cite{lora}, $y=\textbf{W}_{0}x+\Delta\textbf{W}x$, where $\Delta\textbf{W}$ consists of two learnable matrices: $\textbf{B}\in\mathbb{R}^{b \times 4}$, $\textbf{A}\in\mathbb{R}^{4 \times a}$, to fine-tune the pretrained LLARVA with six robotic skills, offering the following insightful observations. 

\textbf{Observation 1: \textit{Skill parameter subspaces and skill semantic spaces have similar correlations}}. One straightforward idea is that if two skills are similar, the two parameter spaces for learning them also have similar correlations. We explore whether the semantic subspace and parameter subspace have similar correlations. Specifically, we calculate six skills' parameter subspace similarity matrices and semantic subspace similarity matrices. The parameter subspace similarity matrices measure the similarity between each of the skill parameter subspaces $\Delta\textbf{W}$. The semantic subspace similarity matrices measure the similarity between each of the skill instruction semantic subspaces $E_{T}(\mathcal{I}_{t})$ (CLIP text encoder $E_{T}(\cdot)$ \cite{CLIP}). We summarize the two similarity matrices in Figure \ref{fig_AF1} (a). We find that the two similarities are consistent in each skill task, which suggests that the instruction semantic space can be used to associate skill parameter subspaces. And in practice, for a new skill, its skill semantic subspace is explicit and easily obtainable, i.e., it can be obtained by encoding user manipulation instructions; while its skill parameter subspace is implicit and requires specific learning for obtaining. Therefore, this observation suggests we can estimate the new skill’s parameter subspace from the similarity of the current semantic subspace with all previous semantic subspaces. 

\textbf{Observation 2: \textit{Skill subspaces required for different skills are distributed across different layers}}. We also explore the distribution of learning skill spaces required for different skills across different transformer layers (from $l$-th to $L$-th layer). Specifically, we calculate the average value of each skill subspace in different layers $\{\Delta\bar{\textbf{W}}_{t}\}^{L}_{l=1}$, as shown in Figure \ref{fig_AF1} (b). Based on this experimental result, we find that the distributions of learnable skill subspaces required for different skills across layers show significant differences, suggesting that the importance of different layers for different skills is also not the same. Therefore, to better learn specific knowledge for different skills, we should not assign the same fixed learning subspace for all manipulation skills. 

\textbf{Observation 3: \textit{Skill-shared and skill-specific knowledge are naturally decoupled in their subspaces}}. We further explore the learning consistency of skill parameter subspaces for robotic manipulation multi-skills. Specifically, we calculate the similarity between each $t$-th skill subspace $\textbf{B}_{t}, \textbf{A}_{t}$ and the other five skills $\{\textbf{B}_{j}, \textbf{A}_{j}\}_{j \neq t}$, separately. We summarize their average values to obtain average similarity for each skill, as in Figure \ref{fig_AF1} (c). We find that the similarity of $\textbf{A}_{t}$ is much greater than the similarity of $\textbf{B}_{t}$, which suggests $\textbf{A}_{t}$ tend to be consistent while $\textbf{B}_{t}$ lack consistency. And a similar pattern was also reported by HydraLoRA \cite{HydraLoRA} in natural language understanding and generation tasks. Therefore, this observation suggests that the skill subspace $\textbf{A}$ tends to learn skill-shared knowledge, the subspace $\textbf{B}$ tends to learn skill-specific knowledge, naturally. 

\section{The Proposed SkillsCrafter}

\begin{figure*}[!t]
\centering
\includegraphics[width=0.99\linewidth]{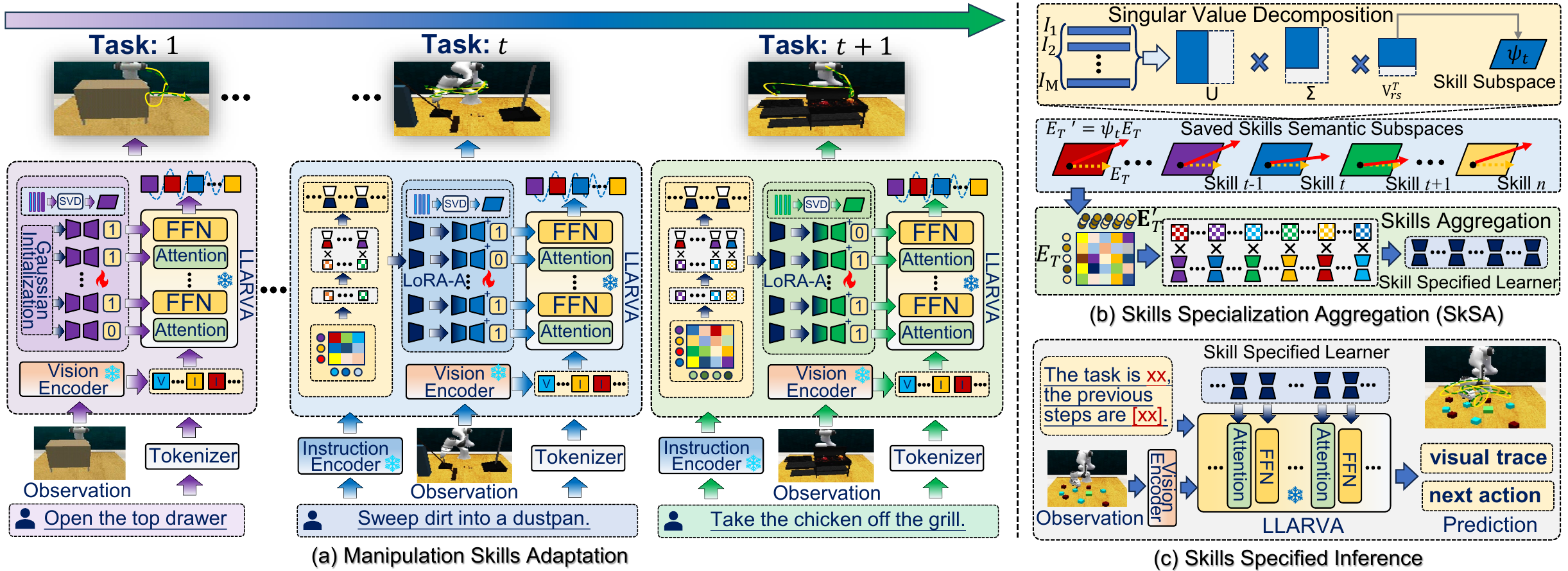}
\caption{Illustration of the proposed SkillsCrafter pipeline. It includes (a) a \textit{Manipulation Skills Adaptation} to achieve retaining the old skills knowledge while inheriting the shared knowledge between new and old skills to facilitate learning of new skills; (b) a \textit{Skills Specialization Aggregation} to compute inter-skills similarity in skill-specific subspaces to achieve adaptive aggregation of skills knowledge; (c) a \textit{Skills Specified Inference} to loads aggregated knowledge to achieve any skill manipulation inference.
}
\label{fig_2}
\vspace{-3mm}
\end{figure*}

The overall pipeline of SkillsCrafter is illustrated in Figure \ref{fig_2}. It includes a Manipulation Skills Adaptation (MSkA) to learn skills continually, a Skills Specialization Aggregation (SkSA) to aggregate learned knowledge, and Skills Specified Inference (SkSI) to load knowledge for skill inference.

\subsection{Manipulation Skills Adaptation} 

As shown in Figure \ref{fig_2}(a), in order to learn the $t$-th skill $S_{t}$, we use the LoRA \cite{lora} to finetune the pretrained LLARVA $\mathcal{F}_{\theta_{0}}$ on skill-specific data $S_{t}=\{\mathcal{O}_{t},\mathcal{I}_{t}\}$, and then obtain an updated model $\mathcal{F'}_{\theta'_{t}}$, where $\theta'_{t}=\theta_{0}+\Delta\theta_{t}$, $\Delta\theta_{t} = {\{\Delta \textbf{W}^{l}_{t}}\}^{L}_{l=1}$, and $\Delta \textbf{W}^{l}_{t}= \textbf{B}^{l}_{t}\textbf{A}^{l}_{t}\in\mathbb{R}^{b_{l} \times a_{l}}$ is the updated low-rank weight in $l$-th layer of a total of $L$ layers. $\textbf{B}^{l}_{t}\in\mathbb{R}^{b_{l} \times r}$ and $\textbf{A}^{l}_{t}\in\mathbb{R}^{r \times a_{l}}$ are the low-rank weights. To address LLCRM, a trivial solution is to train and save low-rank weights for all skills so far, and then load their specific weights at inference time. However, many skills share and have specific knowledge with each other; exploring and exploiting the knowledge promotes better new skills learning. For better skills adaptation, different from traditional LoRA, we use subspace $\textbf{A}_{t}$ to learn skills-shared knowledge and $\textbf{B}_{t}$ to learn skills-specific knowledge, based on \textit{\textbf{Observation 3}}. 

\textbf{Skills-Shared Knowledge:} For skills-shared knowledge, we propose an adapter inheritance strategy to exploit the knowledge learned so far to promote better new skills learning. Specifically, in learning the initial skill $S_{1}$, we randomly initialize $\textbf{A}^{l}_{1}$ and zero initialize the $\textbf{B}^{l}_{1}$ matrix following standard LoRA \cite{lora}. When learning subsequent new skills $S_{t}, t>1$ continually, we perform knowledge aggregation and inheritance on the $\textbf{A}^{l}_{t}$. We compute skills similarity as aggregation weight $\Omega_{t}=\{\omega_{1},\omega_{2},...\omega_{t-1}\}$ within skill semantic subspace $\Psi_{t}$ based on a SkSA Module (as Eq.(\ref{eqw})) to aggregate all the learned adapters $\mathcal{A}_{t} = \{\textbf{A}^{l}_{g}\}_{g=1}^{t-1}$:
\begin{equation}
{{\textbf{A}^{l}_{t*} = \Omega_{t} \odot \mathcal{A}_{t},} \ \ \ \Omega_{t} = SkSA(\mathcal{I}_{t}, \Psi_{t-1}),}
\end{equation}
where $\Psi_{t-1}=\{\psi_{1},\psi_{2},...,\psi_{t-1}\}$ are the saved skills semantic subspaces learned so far, and $\odot$ is the Hadamard product. The aggregated adapter $\textbf{A}^{l}_{t*}$ is used as the initialization of $\textbf{A}^{l}_{t}$ to inherit the skills-shared knowledge. 

\textbf{Skill-Special Knowledge:} To further consolidate the skill shared-knowledge learning on skill parameter subspace $\textbf{A}^{l}_{t}$ and keep the structure of shared knowledge stable, we also perform the skills orthogonal optimization on skill parameter subspace $\textbf{B}^{l}_{t}$. Specifically, during adapting the $t$-th skill, we prefer that the skill-special parameter subspace $\textbf{B}^{l}_{t}$ be orthogonal to the previous skill: $\sum^{t-1}_{i=1}\sum^{L}_{l=1}\operatorname{tr}((\textbf{B}^{l}_{i})^\mathrm{T}\textbf{B}^{l}_{t})=0$. Furthermore, to prevent subspace $\textbf{B}^{l}_{t}$ from degenerating into the trivial optimal solution, i.e., the zero matrix (without learning any knowledge), we perform L2 normalization on the subspace $\textbf{B}^{l}_{t}$, i.e. project them onto the unit sphere: 
\begin{equation}
\small
R_{t}\!\!=\!\!\sum\!\operatorname{tr}((\widetilde{\textbf{B}}^{l}_{i})^\mathrm{T}\widetilde{\textbf{B}}^{l}_{t}), \ 
\widetilde{\textbf{B}}^{l}_{i} \!=\! \frac{\textbf{B}^{l}_{i}}{\mid\mid\textbf{B}^{l}_{i}\mid\mid_{F}\!+\epsilon}, \widetilde{\textbf{B}}^{l}_{t} \!=\! \frac{\textbf{B}^{l}_{t}}{\mid\mid\textbf{B}^{l}_{t}\mid\mid_{F}\!+\epsilon}.
\end{equation}
Thus the $t$-th skill adaptation loss for the LLM-based agent $\mathcal{F}$ performing auto-regressive action generation training is: 
\begin{equation}
{\mathcal{L}_{t} = -\sum_{n=1}^{N}logP_{t}(\hat{A}_{n:n+z-1}, \hat{\mathcal{P}}_{n:N \mid \mathcal{IO}_{t,n}}) + \lambda R_{t},}
\end{equation}
where the $P_{t}(\hat{A}_{n:n+z-1}, \hat{\mathcal{P}}_{n:N \mid \mathcal{I}_{t} \mathcal{O}_{t,n}})$ denotes the predicted probability of ground-truth next $z$ steps actions $\hat{A}_{n:n+z-1}$ and visual traces $\hat{\mathcal{P}}_{n:N}$ under the current observation $\mathcal{IO}_{t,n}=\{\mathcal{I}_{t}, \mathcal{O}_{t,n}\}$, and $z\leq N$. Specifically, the auto-regressive generation predicted probability is calculated:
\begin{equation}
\small
{P_{t}(\hat{A}_{n:n+z-1}, \hat{\mathcal{P}}_{n:N \mid \mathcal{IO}_{t,n}}) = \!\!\!\!\prod_{i=n}^{n+z-1}\!\!P_{t,\Delta\theta_{t}}(x_{i} \mid \mathcal{IO}_{t,n})},
\end{equation}
where $\Delta\theta_{t}$ are the all trainable parameters, and $x_{i}$ is the current prediction token constituting manipulation actions.

In addition, due to the specificity of different skills, static loading of LoRA for fixed layers may lead to overfitting for simple skills (too many parameters) and underfitting for complex tasks (insufficient parameters). Based on \textit{\textbf{Observation 2}}, we propose a dynamic sparse LoRA loading strategy that adaptively decides whether to inject LoRA at each layer using a learnable Gumbel-Softmax gating mechanism. It uses Gumbel noise to approximate discrete choices as continuous distributions for enabling end-to-end training. Specifically, in each layer-wise LoRA $\Delta\textbf{W}^{l}_{t}$, we add a learnable linear layer $H_{t,l}(\cdot)$ for each embedding $em_{t,l}$ to obtain $\mathbf{e}_{t,l}\! =\! H_{t,l}(em_{t,l}) \!\in\! \mathbb{R}^2$, where the two dimensions correspond to the options injection or no injection. We inject Gumbel noise $u_{t,l}^{(i)}\! =\! -\log(-\log(U_{t,l}^{(i)}))$, where $U \!\sim\!\text{Uniform}(0,1)$, compute Gumbel-Softmax distribution:
\begin{equation}
G_{t,l}^{(i)} = \frac{\exp((e_{t,l}^{(i)} + u_{t,l}^{(i)})/\tau_g)}{\sum_{j=1}^2 \exp((e_{t,l}^{(j)} + u_{t,l}^{(j)})/\tau_g)}, 
\end{equation}
where $\tau_g$ is a temperature controlling decision discreteness. We use hard decision $g_t^l \!=\! \mathbb{I}(\arg\max_i G_{t,l}^{(i)} \!=\!2)$ as forward propagation, and a soft gating $g_t^l \! = \!  G_{t,l}^{(2)}\!\in\![0,1]$ as backward propagation for training, to dynamically inject sparse LoRA:
\begin{equation}
\theta'_{t} = \theta_{0} + g_{t}^{l} \cdot \Delta\textbf{W}^{l}_{t}.
\end{equation}
To encourage sparse injection, we add a sparsity regularization term $\mathcal{L}_{s}\!\! = \!\! \sum_{l=1}^L g_t^l$, and jointly optimize loss $\mathcal{L}_{t}\!+\!\lambda_{s}\mathcal{L}_{s}$.

\subsection{Skills Specialization Aggregation}

After learning a series of skills continually, SkillsCrafter requires recalling which of the learned knowledge is valuable for the current skill, and exploiting the knowledge to better complete the current skill. Based on \textit{\textbf{Observation 1}}, we can estimate the similarity of skill parameter subspaces from the similarity of skill semantic subspaces. Thus, we propose the Skills Specialization Aggregation (SkSA) module to adaptively aggregate the already learned adapters $\{\Delta \textbf{W}_{g}\}_{g=1}^{t}$ based on skill semantic subspaces. Specifically, we use Singular Value Decomposition (SVD) \cite{SVD} to obtain a common skill semantic space for skills with the same skill but different instructions. Each skill semantic space is a discriminant subspace corresponding to its skill, and we perform similarity comparisons within these semantic subspaces to obtain skill similarity. The existing knowledge (parameter subspaces) is aggregated based on the similarity to achieve learning or inference of the current skill.

\textbf{SVD for Common Skill Semantic Subspaces:} When adapting new skills, we also calculate and store the common skill semantic spaces from diverse skill instructions. Specifically, as Figure \ref{fig_2}(b), for each $t$-th skill, we use the CLIP text-encoder $E_{T}(\cdot)$ \cite{CLIP} to extract embeddings for diverse user instructions $X_{t} = [E_{T}(\mathcal{I}_{t}^{1}); E_{T}(\mathcal{I}_{t}^{2}); ...; E_{T}(\mathcal{I}_{t}^{M})]\in \mathbb{R}^{M\times d}$, where $M$ is the total number of instruction samples from $t$-th skill $S_{t}$ and $d$ is the extracted embedding dimension. These embeddings are then subjected to SVD to extract skill semantic subspace: 
\begin{equation}
X_{t}=U_{t}\Sigma_{t} V_{t}^{T},
\end{equation}
where $U_{t}\in \mathbb{R}^{M\times M}$ is the left singular vector matrix, $\Sigma_{t}\in \mathbb{R}^{M\times d}$ is the singular values matrix and $V_{t}\in \mathbb{R}^{d\times d}$ is the right singular vector matrix. In order to maximize the distribution of the $t$-th common skill semantic subspace to reflect the distribution of the skill $S_{t}$, we select the first $r_{s}$ rows of the right singular vector matrix $V_{t}$ to represent the distribution of the embeddings of $t$-th skill, denoted $V_{t,r_{s}} =V_{[:,0:r_{s}]}$, thus we store orthogonal projection $\psi_{t}$ for the $S_{t}$ subspace:
\begin{equation}
\psi_{t} = V_{t,r_{s}}V_{t,r_{s}}^{T},
\end{equation}
where $\psi_{t} \in \mathbb{R}^{d\times d}$ is $t$-th skill semantic subspace projection. 

\textbf{Skill Specialization Aggregation:} For new skill adapting or unknown skill inference, we adaptively aggregate all already learned knowledge $\{\Delta \textbf{W}_{g}\}_{g=1}^{t}$ based on skill semantic subspaces $\Psi = \{\psi_{1},\psi_{2},...,\psi_{t}\}$ according to any input instructions $\mathcal{I}_{q}$. Specifically, for skill $S_{q}$, we compute subspace projection $\textbf{E}^{'}(\mathcal{I}_{q})$ of $E_{T}(\mathcal{I}_{q})$ onto every subspace $\Psi$:
\begin{equation}
\textbf{E}^{'}(\mathcal{I}_{q}) = \{\psi_{1}E_{T}(\mathcal{I}_{q}), \psi_{2}E_{T}(\mathcal{I}_{q}), ..., \psi_{t}E_{T}(\mathcal{I}_{q})\},
\end{equation}
where $\textbf{E}^{'}(\mathcal{I}_{q})$ contains the projection of current known or unknown skill $S_{q}$ onto the skill semantic subspaces $\Psi$. Subsequently, we calculate the angle between $E_{T}(\mathcal{I}_{q})$ and every $\textbf{E}_{t}^{'}(\mathcal{I}_{q})$ with cosine similarity, which represents similarity degree of skill $S_{q}$ on subspace $\psi_{t}$ of learned skill $S_{t}$:
\begin{equation}
Sm_{t}(E_{T}(\mathcal{I}_{q}),\textbf{E}_{t}^{'}(\mathcal{I}_{q})) \!=\! \frac{E_{T}(\mathcal{I}_{q}) \textbf{E}_{t}^{'}(\mathcal{I}_{q})}{\parallel E_{T}(\mathcal{I}_{q}) \parallel \cdot \parallel \textbf{E}_{t}^{'}(\mathcal{I}_{q}) \parallel},
\end{equation}
then, each cosine similarity $Sm_{t}$ is transformed exponentially $(Sm_{t})^{\gamma}$ and together constitutes the $N$ weights $\Omega_q$:
\begin{equation} \label{eqw}
\small
\Omega_{q}\!=\!\{\frac{(Sm_{1})^{\gamma}}{\sum^{N}_{i=1} (Sm_{i})^{\gamma}},\! \frac{(Sm_{2})^{\gamma}}{\sum^{N}_{i=1} (Sm_{i})^{\gamma}},\! ...,\! \frac{(Sm_{N})^{\gamma}}{\sum^{N}_{i=1}(Sm_{i})^{\gamma}}\},
\end{equation}
then we use $\Omega_{q}$ as weight to adaptively aggregate all the learned adapters $\{\Delta \textbf{W}^{l}_{t}\}^{N}_{t=1}$ for obtaining aggregated $\Delta \tilde{\textbf{W}}^{l}_{q}$:
\begin{equation} \label{eqa}
\Delta \tilde{\textbf{W}}^{l}_{q} = \Omega_{q} \odot \{\Delta \textbf{W}^{l}_{t}\}^{N}_{t=1}.
\end{equation}

The aggregated $\Delta \tilde{\textbf{W}}^{l}_{q}$ adaptively aggregates previous similar knowledge to efficiently perform the current skill $S_{q}$.

\subsection{Skills Specified Inference}

After employing Eq.(\ref{eqa}) to aggregate all low-rank adapters learned so far, SkillsCrafter loads the aggregated adapter to obtain an updated agent $\mathcal{F'}_{\theta'_{q}}$ specific to the current skill $S_{q}$. As Figure \ref{fig_2}(c), SkillsCrafter loads the aggregated adapter to leverage learned knowledge, which can effectively mitigate the catastrophic forgetting of learned knowledge to better perform any known and unknown skill in the open world. 

\begin{figure}[!t]
\centering
\includegraphics[width=0.95\linewidth]{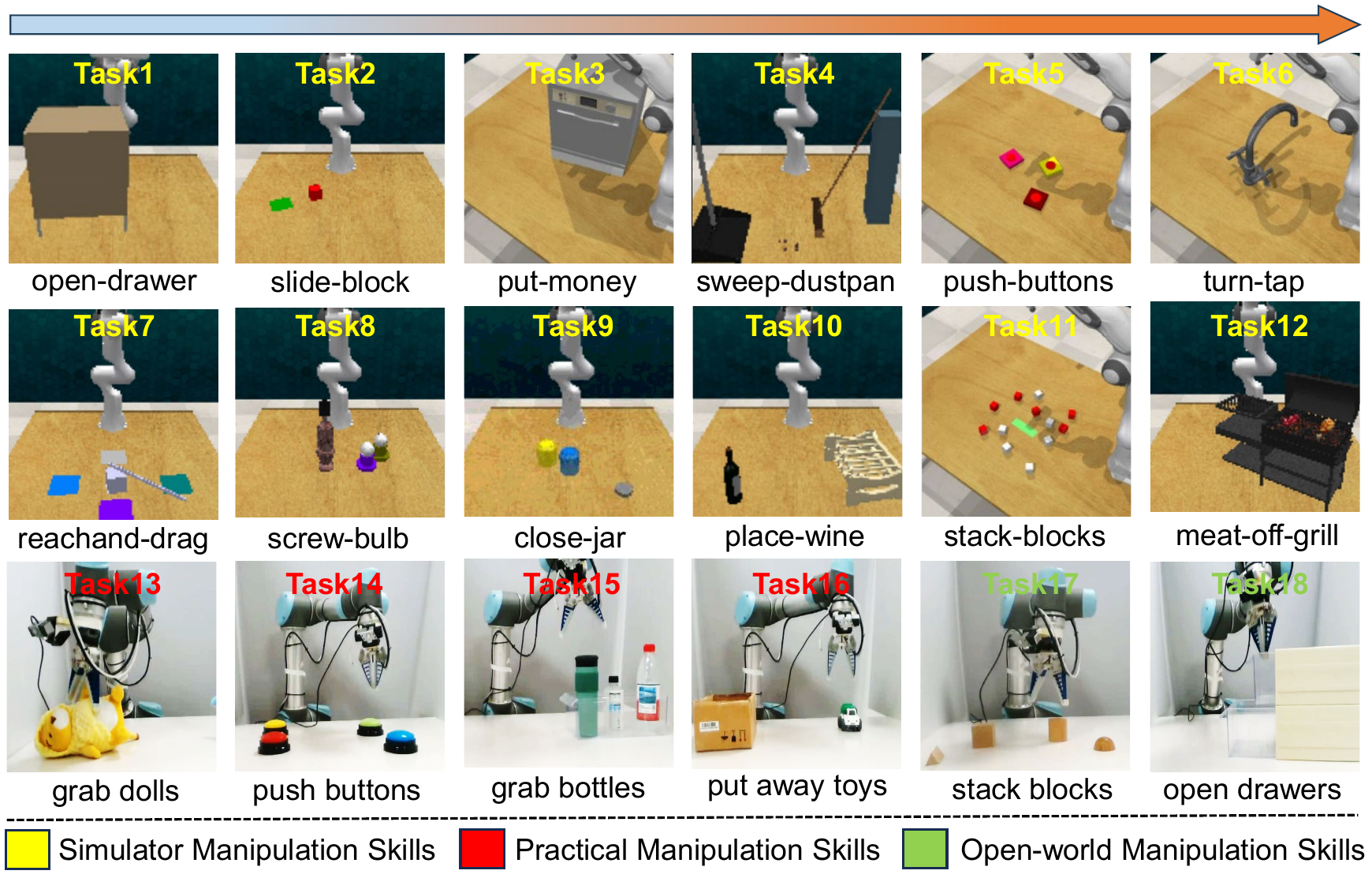}
\caption{Illustration of the experiment robotic skill tasks setting. We establish a total of 12 robotic simulator skills and 6 real-environment robotic skills for incremental learning.
}
\label{fig_4}
\vspace{-3mm}
\end{figure}

\section{Experiments}

\subsection{Implementation Details}

\begin{table*}[t]
\centering
\setlength{\tabcolsep}{1mm}
\renewcommand{\arraystretch}{1}
\setlength{\tabcolsep}{0.86mm} 
\begin{tabular}{l|cccccccccccccccccc|c}

\toprule
Methods & S1 & S2 & S3 & S4 & S5 & S6 & S7 & S8 & S9 & S10 & S11 & S12 & S13 & S14  & S15 & S16 & S17 & S18 & ~Avg.~ \\
\midrule
Seq-FT (sequence fine-tune all) & 0 & 0 & 0 & 0 & 0 & 0 & 0 & 0 & 0 & 0 & 0 & 0 & 12 & 4 & 4 & 80 & 8 & 8 & 6.4 \\

LwF-LoRA \cite{R30} & 0 & 4 & 0 & 4 & 0 & 4 & 0 & 0 & 0 & 0 & 0 & 0 & 8 & 12 & 0 & 84 & 4 & 0 & 6.7 \\

EWC-LoRA \cite{EWC-LoRA} & 0 & 0 & 0 & 0 & 4 & 0 & 0 & 0 & 0 & 0 & 0 & 0 & 16 & 8 & 4 & 80 & 8 & 4 & 6.9 \\

Dense MoLE \cite{Dense} & 4 & 24 & 4 & 12 & 8 & 8 & 8 & 0 & 4 & 0 & 0 & 8 & 24 & 20 & 28 & 80 & 8 & 8 & 13.8 \\

Sparse MoLE \cite{Sparse} & 16 & 32 & 8 & 8 & 12 & 4 & 8 & 4 & 4 & 8 & 0 & 4 & 28 & 16 & 24 & 76 & 4 & 4 & 14.4 \\

MoLA \cite{MoLA} & 12 & 20 & 12 & 8 & 16 & 8 & 12 & 4 & 0 & 4 & 4 & 12 & 44 & 24 & 24 & 80 & 0 & 8 & 16.2 \\

HydraLoRA \cite{HydraLoRA} & 12 & 24 & 16 & 4 & 12 & 4 & 4 & 4 & 4 & 8 & 4 & 8 & 32 & 28 & 28 & 80 & 8 & 16 & 16.4 \\

BranchLoRA \cite{BranchLoRA} & 16 & 28 & 8 & 12 & 8 & 12 & 16 & 0 & 8 & 4 & 0 & 8 & 40 & 36 & 32 & 76 & 12 & 12 & 18.2 \\

O-LoRA \cite{OLoRA} + SkSA & 56 & 72 & \textbf{40} & 60 & 48 & 52 & 36 & 4 & 12 & 8 & 4 & \textbf{76} & 84 & 84 & 76 & \textbf{84} & 36 & 52 & 49.1 \\

SD-LoRA \cite{SD-LoRA} + SkSA & 52 & 80 & 32 & 64 & 44 & 44 & \textbf{48} & 4 & 20 & \textbf{16} & 8 & 72 & 88 & 84 & 80 & 80 & 36 & 48 & 50.0 \\

\midrule

\textbf{SkillsCrafter (ours)}  & \textbf{56} & \textbf{80} & 36 & \textbf{64} & \textbf{48} & \textbf{52} & 44 & \textbf{12} & \textbf{20} & 12 & \textbf{12} & 72 & \textbf{92} & \textbf{84} & \textbf{80} & 80 & \textbf{40} & \textbf{52} & \textbf{52.0} \\
\midrule
\end{tabular}
\vspace{-3mm}
\caption{Test results (Skill-Wise Average Success Rate, ASR $\uparrow$, \%) of comparison experiment with LLCRM settings.}
\label{TAB1}
\end{table*}

\begin{table*}[t!]
\centering
\setlength{\tabcolsep}{1mm}
\renewcommand{\arraystretch}{1}
\vspace{-2mm}
\setlength{\tabcolsep}{1.15mm} 
\begin{tabular}{l|cccccccccccccccccc|c}

\toprule
Methods & S1 & S2 & S3 & S4 & S5 & S6 & S7 & S8 & S9 & S10 & S11 & S12 & S13 & S14  & S15 & S16 & S17 & S18 & ~Avg.~ \\
\midrule
Seq-FT & 100 & 100 & 100 & 100 & 100 & 100 & 100 & 100 & 100 & 100 & 100 & 100 & 88 & 96 & 95 & 5  & 85 & 88 & 92.0 \\

LwF-LoRA  & 100 & 95 & 100 & 95 & 100 & 93 & 100 & 100 & 100 & 100 & 100 & 100 & 92 & 87 & 100 & 0  & 92 & 100 & 91.9 \\

EWC-LoRA & 100 & 100 & 100 & 100 & 93 & 100 & 100 & 100 & 100 & 100 & 100 & 100 & 83 & 91 & 95 & 0  & 85 & 94 & 91.2 \\

Dense MoLE  & 93 & 71 & 91 & 86 & 86 & 86 & 85 & 100 & 86 & 100 & 100 & 90 & 75 & 78 & 68 & 0 & 85 & 88 & 81.5 \\

Sparse MoLE   & 73 & 62 & 82 & 90 & 79 & 93 & 85 & 80 & 86 & 50 & 100 & 95 & 71 & 83 & 73 & 0  & 92 & 94 & 77.0 \\

MoLA  & 80 & 76 & 73 & 90 & 71 & 86 & 77 & 80 & 100 & 75 & 75 & 85 & 54 & 74 & 73 & 0  & 100 & 88 & 75.4 \\

HydraLoRA  & 80 & 71 & 64 & 95 & 79 & 93 & 92 & 80 & 86 & 50 & 75 & 90 & 67 & 70 & 68 & -5  & 85 & 76 & 73.1   \\

BranchLoRA  & 73 & 67 & 82 & 86 & 86 & 79 & 69 & 100 & 71 & 75 & 100 & 90 & 58 & 61 & 64 & 0  & 77 & 82 & 73.3 \\

O-LoRA + SkSA & 7 & 14 & \textbf{9} & 29 & 14 & 7 & 31 & 80 & 57 & 50 & 75 & \textbf{5} & 13 & 9 & 14 & 0  & 31 & 24 & 25.9 \\

SD-LoRA + SkSA  & 13 & 5 & 27 & 24 & 21 & 21 & \textbf{8} & 80 & 29 & \textbf{0} & 50 & 10 & 8 & 9 & 9 & 0 & 31 & 29 & 20.8 \\

\midrule

\textbf{SkillsCrafter}  & \textbf{7} & \textbf{5} & 18 & \textbf{24} & \textbf{14} & \textbf{7} & 15 & \textbf{40} & \textbf{29} & 25 & \textbf{25} & 10 & \textbf{4} & \textbf{9} & \textbf{9} & 0 & \textbf{23} & \textbf{24} & \textbf{16.0} \\
\midrule
\end{tabular}
\vspace{-2mm}
\caption{Test results (Skill-Wise Forgetting Rate, FR $\downarrow$, \%) of comparison experiment with LLCRM settings.}
\label{TAB2}
\vspace{-4mm}
\end{table*}

\noindent \textbf{Manipulation Skills Benchmark Settings:} To validate the proposed  LLCRM task, we construct a comprehensive lifelong robotic manipulation benchmark. The benchmark settings are shown in Figure \ref{fig_4}, including 12 common robotic manipulation skills based on RLBench simulator \cite{RLBench} and 6 real-world manipulation skills. The first 16 skills serve as training and testing tasks for continual learning, while the last two skills are not trained and serve as open-world generalization testing. And our real-world manipulation platform is based on the UR-5 robotic arm with RGB camera. The agent is required to learn the 16 tasks in sequence, and then take a test on each skill after completing all the learning. The id $t$ is agnostic during testing phase. 

\noindent \textbf{Training and Evaluation:}
For fair comparisons, our model and all comparison methods utilize LLARVA \cite{llava} as the baseline backbone. We use the Adam optimizer with an initial learning rate of $1.0 \times 10^{-4}$ for training. All other hyperparameters are consistent with the LLARVA. We perform training and testing using PyTorch 2.1.2 with cu121 on eight NVIDIA RTX 6000 Ada Generation GPUs. Following the previous evaluation methods \cite{llava, R39, RLBench}, we perform each manipulation skill with 25 episodes, and we test the Average Success Rate (ASR) of each skill task:
\begin{equation}
\setlength\abovedisplayskip{1pt}
\setlength\belowdisplayskip{1pt}
ASR_{t} = \frac{1}{25}\sum_{n=1}^{25}SR_{t,n},
\end{equation}
where $SR_{t}$ represents $t$-th skill completion status, $SR_{t}$ is 1 when manipulation $t$-th skill is successfully executed, else, $SR_{t}$ is 0. To evaluate agent continual learning ability, we also use $FR_{t}$ to measure the $t$-th skill forgetting rate:
\begin{equation}
\setlength\abovedisplayskip{1pt}
\setlength\belowdisplayskip{1pt}
FR_{t} = \frac{\frac{1}{25}\sum_{n=1}^{25}SR_{gt,n}-ASR_{t}}{\frac{1}{25}\sum_{n=1}^{25}SR_{gt,n}},
\end{equation}
where $SR_{gt}$ represents the $t$-th skill completion status with learning only the $1$-st to $t$-th skills. And $SR_{gt}$ is 1 when $t$-th skill is successfully executed, else, $SR_{gt}$ is 0. The larger $FR_{t}$, the greater the degree of $t$-th skill forgetting. 

\begin{figure}[!t]
\centering
\includegraphics[width=1\linewidth]{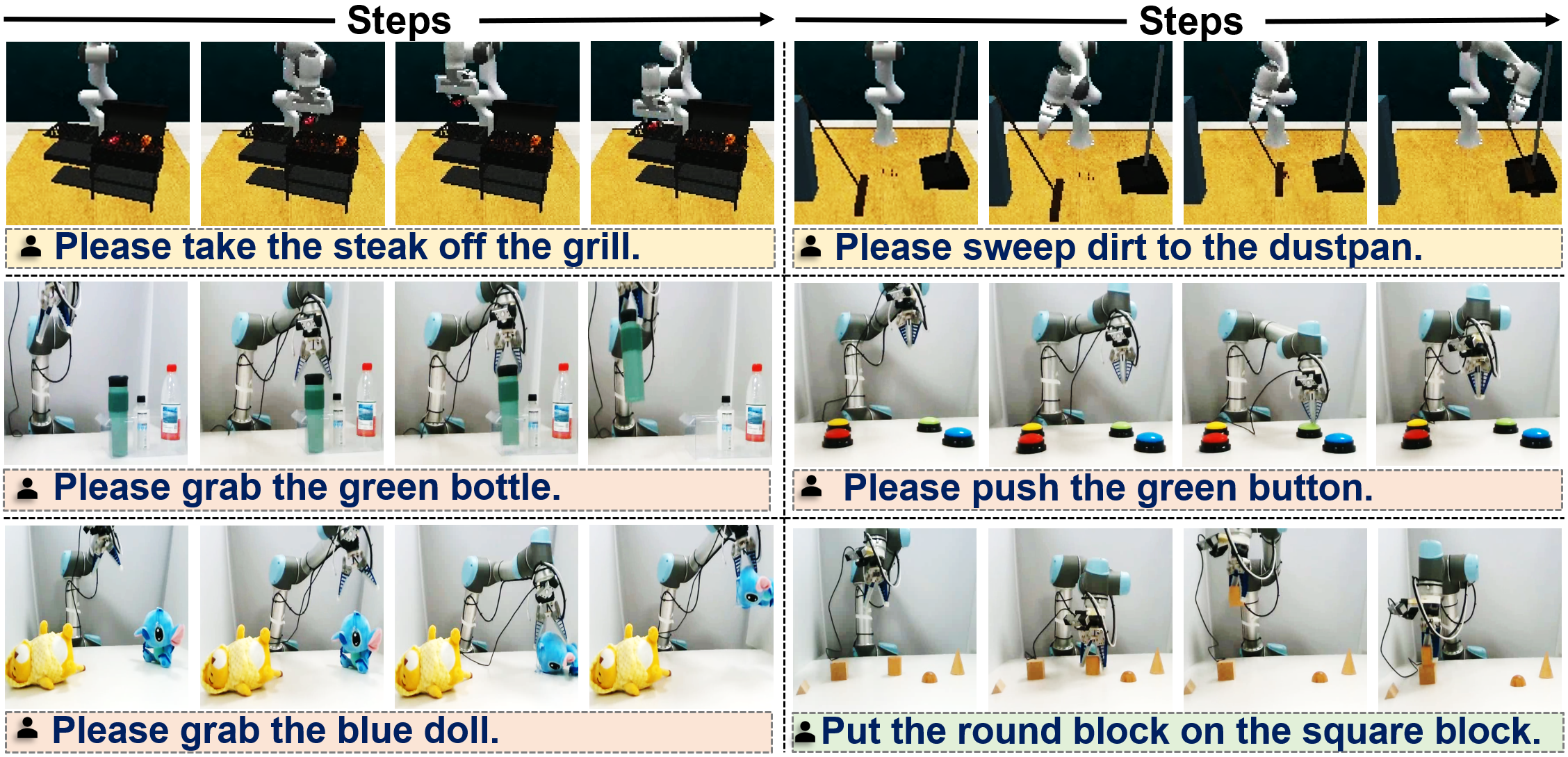}
\caption{Illustration for some visualization examples.}
\label{fig_5}
\end{figure}

\subsection{Comparison Experiment Results}
 
This experiment demonstrates the superior robotic manipulation performance of our SkillsCrafter. Our comparison methods include the SOTA LoRA-based continual learning methods: \textbf{Seq-FT} is fine-tuned for all skills sequences; \textbf{LwF-LoRA} \cite{R30} uses knowledge distillation to retain previous skills performance; \textbf{EWC-LoRA} \cite{EWC-LoRA} penalizes changes to critical parameters of previous skills to mitigate forgetting; \textbf{Dense MoLE} \cite{Dense} uses dense expert routing, while \textbf{Sparse MoLE} \cite{Sparse} uses sparse expert routing in MoE-LoRA; \textbf{MoLA} \cite{MoLA} introduces more experts at a deeper level based on Sparse MoLE; \textbf{O-LoRA} \cite{OLoRA} proposes orthogonal loss to learn task-specific knowledge; \textbf{HydraLoRA} \cite{HydraLoRA} uses $\textbf{A}$ to learn shared knowledge, and multiple $\textbf{B}$ learn specific knowledge, \textbf{BranchLoRA} \cite{BranchLoRA} further enhances the sparse selection mechanism; \textbf{SD-LoRA} \cite{SD-LoRA} performs dynamic combination of previously learned skills LoRA. 
A summary of the evaluation results is shown in Table \ref{TAB1} and Table \ref{TAB2}. Our SkillsCrafter achieves superior manipulation performance with an improved manipulation success rate (ASR: 52.0\%, 2.0\% improvement) and a lower forgetting rate (FR: 16.0\%, 4.8\% reduction). It also demonstrates superior performance on unknown skills (S17, S18), indicating better skill generalization. Some examples of the SkillsCrafter manipulation are provided in Figure \ref{fig_5}.

\subsection{Ablation Studies}

\begin{table}[t]
\centering
\setlength{\tabcolsep}{1mm}
\renewcommand{\arraystretch}{1}
\resizebox{\linewidth}{!}{
\begin{tabular}{l|ccc|cc}
\toprule
{Methods} & GGM & SOT & INA & ~ASR \%~ & ~FR \%~ \\
\midrule
Baseline & \xmarkg & \xmarkg & \xmarkg & 5.8 & 92.8 \\
SkillsCrafter w/o GGM & \xmarkg & \cmark & \cmark & 46.9 & 26.5 \\
SkillsCrafter w/o INA & \cmark & \cmark & \xmarkg & 48.2 & 23.9 \\
SkillsCrafter w/o SOT & \cmark & \xmarkg & \cmark & 49.1 & 22.9 \\
\midrule
\textbf{SkillsCrafter} & \cmark & \cmark & \cmark & \textbf{52.0} & \textbf{16.0}  \\
\midrule
\end{tabular}}
\caption{Ablation study for skills shared/special knowledge. }
\label{ablation1}
\end{table}

\begin{table}[t]
\centering
\setlength{\tabcolsep}{1mm}
\renewcommand{\arraystretch}{1}

\resizebox{\linewidth}{!}
{
\begin{tabular}{l|cccc|cc}
\toprule
Methods & IM & VM & AVG & TOP & ASR~(\%)~$\uparrow$ & FR~(\%) \\
\midrule
SkillsCrafter w/ TOP & \xmarkg & \xmarkg & \xmarkg & \cmark  & 50.4 & 16.5 \\
SkillsCrafter w/ Avg & \xmarkg & \xmarkg & \cmark  & \xmarkg & 44.4 & 26.3 \\
SkillsCrafter w/ VM  & \xmarkg & \cmark  & \xmarkg & \xmarkg & 48.9 & 20.2 \\
\midrule
\textbf{SkillsCrafter} & \cmark & \xmarkg & \xmarkg & \xmarkg & \textbf{52.0} & \textbf{16.0}  \\
\midrule
\end{tabular}
}
\caption{Ablation study on skill knowledge aggregation.}
\label{ablation2}
\end{table}

\noindent \textbf{How Do Skills Shared and Special Knowledge Perform?} These ablation studies' results about skills shared and special knowledge are summarized in Table \ref{ablation1}. Specifically, the “GGM” denotes the Gumbel-softmax Gating Mechanism based sparse loading strategy, which enhances the exploration of the skill-special knowledge; The “INA” denotes the Inheritance $\textbf{A}$ strategy, which enhances the exploration of the skill-shared knowledge; The “SOT” denotes the Skills Orthogonal Term, which also enhances the exploration of the skill-special knowledge and consolidates the structure of LoRA's skill-shared and skill-special knowledge. Based on the ablation results, the proposed methods enable SkillsCrafter to explore and exploit skill-shared and special knowledge between diverse skills, effectively reducing catastrophic forgetting and improving new skill learning.

\noindent \textbf{How Do Skills Knowledge Aggregation Perform?} The ablation studies results about the knowledge aggregation methods are summarized in Table \ref{ablation2}. Specifically, the “IM” denotes the Knowledge aggregation based on Instruction Matching; The “VM” denotes the Knowledge aggregation based on Vision Matching; The “Avg” denotes the average pooling for all instruction encodings, rather than using SVD; The “TOP” denotes the selection of the most similar LoRA in SkSA. Based on the ablation results, the proposed aggregation methods enable SkillsCrafter to exploit learned skill knowledge, effectively improving the LLCRM performance. 

\section{Related Works}

\textbf{Language-Conditioned Robotic Manipulation.} Significant progress has been made in robotic manipulation, enabling agents to tackle increasingly complex tasks \cite{R6, R8, R20, R37, R14, Uniyolo, ICIRA1, NBAgent, seqwalker, PixelVLA}. For example, VIMA \cite{R24} introduces a multi-modal prompting framework that formulates diverse manipulation tasks as a sequence modeling problem. Some methods \cite{R23, R17, CASM} use voxelized 3D point cloud representations to better support sophisticated manipulations. PerAct \cite{R39} improves manipulation performance by converting RGB-D observations into voxel grids and discretizing the action space. GNFactor \cite{R53} improves generalization by training the rendering of neural radiance fields. Despite advances, allowing robots to continually acquire skills without forgetting learned skills remains a challenge.

\noindent \textbf{Continual Learning.} It is essential for building adaptive AI systems that evolve over time \cite{R46}. Existing methods can be divided into three categories. Parameter regularization methods \cite{R33, R30, R9, cite5} constrain weight updates to preserve prior knowledge. Architecture-based methods \cite{CRISP, R25, R48, R45, CCVD} dedicate specific network components to different tasks. Replay-based methods use either stored data \cite{R4, R41} or generated samples \cite{R29, R49, R40} to mitigate forgetting. Several works explore applying continual learning in robotics \cite{R3, R15, R18, WXDICLR1, WXDICLR2}. For example, LOTUS \cite{R42} enables continual learning by gradually expanding a skill library with new task demonstrations. However, these methods do not focus on skill-shared knowledge for better skills learning. 

\section{Conclusion}

In this paper, we propose \textit{SkillsCrafter}, a novel language-conditioned robotic manipulation model designed to continually learn multiple manipulation skills while reducing catastrophic forgetting of old skills. To explore shared and specific knowledge across multiple skills, we propose a \textit{Manipulation Skills Adaptation} to inherit the skill knowledge between new and old skills to facilitate learning of new skills. We propose \textit{Skills Specialization Aggregation} from the skill semantic space aggregation parameter space to achieve inference of any skills. Experiments in both simulation and real-world environments confirm its effectiveness.

\section{Acknowledgments}
This work was supported by the National Key Research and Development Program of China under Grant 2024YFB4707700, the National Natural Science Foundation of China under Grant T2596040, T2596045 and U23A20343, CAS Project for Young Scientists in Basic Research, Grant YSBR-041, Liaoning Provincial ``Selecting the Best Candidates by Opening Competition Mechanism" Science and Technology Program under Grant 2023JH1/10400045, Fundamental Research Project of SIA under Grant 2024JC3K01.

\bibliography{aaai2026}

\newpage

\section{Notations and Definitions}\label{IN}

For the convenience of readers, we summarize some important notations and definitions used in this paper in Table \ref{notation}.

\begin{table}[!ht]
\renewcommand{\arraystretch}{1.1}
    \centering
    \scalebox{1}{
        \begin{tabular}{c|l}
            \toprule
            \textbf{Notations} & \textbf{Definitions} \\
            \midrule
            $\mathcal{I}$ & user language instruction  \\
            $\mathcal{O}$ & vision observation \\
            $\mathcal{V}(\mathcal{O})$ & CLIP vision-encoder \\
            $\mathcal{S}$ & multi-skills sequence set \\
            $\{S_{1}, S_{2}, ..., S_{t-1}\}$ & old skills \\
            $S_{t}$ & current skill \\
            $l$ & $l$-th transformer layer \\
            $L$ & A total of $L$ layers of transformer \\
            $r$ & rank of LoRA \\ 
            $\Delta \theta_t$ & all trainable parameters \\
            $\mathcal{\textbf{A}}$ & low-rank weight A \\
            $\mathcal{\textbf{B}}$ & low-rank weight B \\
            $\Delta \textbf{W}$ & skill parameter subspace  \\
            $\mathcal{F}_{\theta_{0}}$ & pretrained LLARVA  \\
            $\mathcal{F'}_{\theta'_{t}}$ & an updated model with $t$-th skill\\
            $E_T(\cdot)$ & CLIP text encoder  \\
            $\Omega_t$ & aggregation weight with $t$-th skill \\
            $\Psi_t$ & skill semantic subspace \\
            $H_{t,l}(\cdot)$ & learnable linear layer \\
            $\mathbf{e}_{t,l}$ & logits function \\
            $g_t^l$ & dynamically decision loading \\
            $X$ & instruction samples \\
            $U$ & left singular vector matrix \\
            $V$ & right singular vector matrix \\
            $Sm(\cdot)$ & cosine similarity \\
            $\mathbb{I}(\cdot)$ & indicator function \\
            \bottomrule
        \end{tabular}
    }
\caption{Some important notations and definitions.}
    \label{notation}
\end{table}

\section{Comparisons Implementation}\label{experiment}

We provide a detailed experimental implementation. For fair comparisons, our model and all comparison methods utilize LLARVA \cite{llava} as the baseline backbone. We use the Adam optimizer with an initial learning rate of $1.0 \times 10^{-4}$ for training. All the other hyperparameters are consistent with the LLARVA, and we summarize the used hyperparameters in Table \ref{tab:hyperparameters}. We perform training and testing using PyTorch 2.1.2 with cu121 on eight NVIDIA RTX 6000 Ada Generation GPUs (system Ubuntu 22.04.3). Following the previous evaluation methods \cite{llava, R39, RLBench}, we perform each manipulation skill with 25 episodes, and we test the Average Success Rate (ASR) and skill Forgetting Rate (FR) of each manipulation skill task. 

\begin{table}[t]
\renewcommand{\arraystretch}{1.1}
\begin{center}
\begin{tabular}{cc}
\toprule
\textbf{Hyperparameters Name} & \textbf{Value} \\
\hline
image size & 128 × 128 × 3 \\
per device train batch size $B$ & 16 \\
learning rate & 0.0001 \\
num train epochs & 4 \\
computational bits & 16 \\
eval max steps & 150 \\
mm vision select layer & -2 \\
rank of LoRA $r$ & 8\\
inference episodes & 5\\
transformation term $\gamma$ & 5\\
temperature parameter $\tau_g$ & 1 \\
$r_s$ of the right singular vector matrix & 20 \\
$d$ of projection $\psi_{t} \in \mathbb{R}^{d\times d}$ & 512 \\
$\lambda$ of orthogonal regularizer & 0.1 \\
$\lambda_{s}$ of sparsity regularizer & 0.01 \\\
perturbation $\epsilon$  & 0.01 \\

\bottomrule
\end{tabular}
\end{center}
\caption{Hyper-parameters used in SkillsCrafter.}
\label{tab:hyperparameters}
\end{table}

We compare SkillsCrafter against the following state-of-the‑art LoRA‑based continual learning baselines: Sequential fine‑tuning (Seq‑FT) simply continues to fine‑tune the shared pre‑trained model on each new skill in turn, without any explicit mechanism to prevent forgetting. While straightforward, it typically suffers from catastrophic forgetting as new updates overwrite parameters important to earlier tasks. Learning without Forgetting-LoRA (LwF-LoRA) \cite{R30} augments LoRA by applying knowledge‑distillation: when training on a new skill, the current model is encouraged to match the soft logits of the previous‑skill model on a held‑out set. This preserves performance on past tasks by penalizing divergence from earlier outputs. Elastic Weight Consolidation-LoRA (EWC-LoRA) \cite{EWC-LoRA} adapts the Fisher information to LoRA by adding a quadratic penalty on changes to parameters deemed important for prior skills. The importance weights are computed via the Fisher information matrix, discouraging updates that disrupt performance on earlier tasks. The Mixture‑of‑LoRA‑Experts (MoLE) framework attaches multiple LoRA adapters (“experts”) and routes each input to all experts (Dense MoLE) \cite{Dense} or to a learned sparse subset (Sparse MoLE) \cite{Sparse}. Dense routing ensures maximal capacity but at a higher computation cost; sparse routing reduces overhead by selecting only the top‑$k$ experts per instance ($N=16, k=2$). Building on Sparse MoLE, Mixture‑of‑LoRA‑Adapters (MoLA) \cite{MoLA} increases the depth and number of experts ($N=32$). Orthogonal‑LoRA (O-LoRA) \cite{OLoRA} introduces an orthogonality constraint on the LoRA update matrices: for each task, the new adapter’s subspace is enforced to be orthogonal to the span of all previous adapters. This explicit decorrelation reduces interference but can limit positive transfer when tasks are related. HydraLoRA \cite{HydraLoRA} decomposes each task’s LoRA update into a shared component $\mathbf{A}$ and a task‑specific component $\mathbf{B}$. The shared $\mathbf{A}$ captures common knowledge across tasks, while distinct $\mathbf{B}$’s focus on individual skill nuances. Extending HydraLoRA’s idea of split updates, BranchLoRA \cite{BranchLoRA} incorporates a learned sparse selection tree: at each layer, a lightweight router chooses a branch among multiple $\mathbf{B}$ modules, effectively growing a task hierarchy that adapts granularity based on skill similarity and complexity. SD‑LoRA \cite{SD-LoRA} performs on‑the‑fly linear combinations of all previously learned LoRA adapters. Please note that both O-LoRA and SD-LoRA train multiple sets of LoRA and select the LoRA with the highest similarity based on our proposed semantic similarity matching strategy during inference. To ensure fairness in the comparison, these MoE-LoRA methods, i.e., MoLE, HydraLoRA and BranchLoRA use 16 experts (number of tasks).

\section{Algorithm Summary}\label{ASS}

For ease of readers understanding, a summary of the proposed SkillsCrafter learning algorithm is provided in the Algorithm \ref{alg:training_process}, and a summary of inference algorithm is provided in the Algorithm \ref{alg:training_process2}.

\begin{algorithm}[t!]
\setlength{\abovecaptionskip}{1cm}
    \caption{The training pipeline of SkillsCrafter}
    \label{alg:training_process}
    \begin{algorithmic}[1]
        \STATE Prepare stream data $\mathcal{S}=\{S_{1}, S_{2}, ..., S_{T}\}$, where $S_{t}=\{\mathcal{O}_{t},\mathcal{I}_{t}\}$. Learnable parameters: $\Delta \textbf{W}_{t}= \textbf{B}_{t}\textbf{A}_{t}\in\mathbb{R}^{b \times a}$.
        
        \FOR{$t=0$ \textbf{to} $t=T$ in $\mathcal{S}_{t}$}
            \IF{$t=1$}
            \STATE Kaiming initialize $\textbf{A}_{t}$ and zero initialize the $\textbf{B}_{t}$.
            \ELSE
            \STATE Use text-encoder to extract embeddings $E_{T}(\mathcal{I}_{t})$.
            \STATE Compute subspace projection $\textbf{E}^{'}(\mathcal{I}_{t})$ with Eq.(9).
            \STATE Compute aggregated weight $\Omega_{t}$ with Eq.(10)(11).
            \STATE Aggregate all the learned adapters  $\textbf{A}_{t*}$ with Eq.(1). 
            
            \STATE Perform $\textbf{A}_{t*}$ initialize $\textbf{A}_{t}$ and zero initialize the $\textbf{B}_{t}$.
            \ENDIF
        
            \FOR {$(\mathcal{O}^{i}_{t},\mathcal{I}^{i}_{t})\in\mathcal{S}_{t}$}

            \STATE Compute skills orthogonal term $R_{t}$ with Eq.(2). 
            \STATE Compute sparsity orthogonal term $\mathcal{L}_{s}$. 
            \STATE Compute AR-generation loss $\mathcal{L}_{t}$ with Eq.(3). 
            \STATE Optimize joint loss $\mathcal{L}_{t}\!+\!\lambda_{s}\mathcal{L}_{s}$ to update parameters. 
            
            \ENDFOR 

            \STATE Use SVD to calculate common skill semantic subspace from diverse skill instructions with Eq.(7)(8). 

            \STATE Save $t$-th LoRA $\textbf{B}_{t}\textbf{A}_{t}$ and subspace projection $\psi_{t}$.
            
        \ENDFOR
        
    \end{algorithmic}
\end{algorithm}

\begin{algorithm}[t!]
\setlength{\abovecaptionskip}{1cm}
    \caption{The inference pipeline of SkillsCrafter}
    \label{alg:training_process2}
    \begin{algorithmic}[1]
        \STATE For any known or unknown skill $S_{q}$ in the open-world.
        
            \FOR {$(\mathcal{O}^{i}_{q},\mathcal{I}^{i}_{q})\in\mathcal{S}_{q}$}

            \STATE Use text-encoder to extract embeddings $E_{T}(\mathcal{I}_{q})$.
            \STATE Compute subspace projection $\textbf{E}^{'}(\mathcal{I}_{q})$ with Eq.(9).
            \STATE Compute aggregated weight $\Omega_{q}$ with Eq.(10)(11).
            \STATE Aggregate all learned adapters  $\textbf{A}_{q}\textbf{B}_{q}$ with Eq.(12). 

            \STATE Load the aggregated $\textbf{A}_{q}\textbf{B}_{q}$ with $H_{q}(\cdot)$ corresponding Top-1 similarity with learned skills for $S_{q}$ inference.
            
            \ENDFOR 

    \end{algorithmic}
\end{algorithm}

\begin{table*}[t]
\centering
\setlength{\tabcolsep}{1.4mm}
\renewcommand{\arraystretch}{1.3}
\small  

\begin{tabular}{c c c}
\hline
\textbf{No.} & \textbf{Skill in Simulator} & \textbf{Instruction Examples} \\
\hline
1 & open-drawer & ``open the top drawer", ``open the middle drawer", ``open the bottom drawer" \\
2 & slide-block & ``slide the block to target" \\
3 & put-money & ``put the money away in the safe on the top shelf", ``put the money away in the safe on the bottom shelf" \\
4 & sweep-dustpan & ``sweep dirt to dustpan" \\
5 & push-buttons & ``push the maroon button", ``push the maroon button, then push the blue button", ``push the purple button" \\
6 & turn-tap & ``turn right tap", ``turn left tap" \\
7 & reach-and-drag & ``use the stick to drag the cube onto the gray target" \\
8 & screw-bulb & ``screw in the azure light bulb", ``screw in the silver light bulb", ``screw in the purple light bulb" \\
9 & close-jar & ``close the blue jar", ``close the lime jar", ``close the yellow jar", ``close the magenta jar" \\
10 & place-wine & ``stack the wine bottle to the left of the rack", ``stack the wine bottle to the right of the rack" \\
11 & stack-blocks & ``stack 2 magenta blocks", ``stack 3 navy blocks", ``stack 4 violet blocks", ``stack 4 olive blocks" \\
12 & meat-off-grill & ``take the steak off the grill", ``take the chicken off the grill" \\
\hline
\textbf{No.} & \textbf{Skill in Real-world} & \textbf{Instruction Examples} \\
\hline
13 & grab dolls & ``grab the yellow doll", ``grab the blue doll", ``grab the black doll", ``grab the red doll" \\

14 & push buttons & ``push the blue button, then push the yellow button", ``push the green button, then push the red button" \\

15 & grab bottles& ``grab the green bottle", ``grab the bottle with the white cap", ``grab the bottle with the red cap" \\

16 & put away toys & ``put the toy car into the box", ``put the steel ring into the box", ``put the toy duck into the box" \\

17 & stack blocks & ``put the round wooden block onto the square block", ``put the square wooden block onto the round block" \\

18 & open drawers & ``close the top drawer", ``open the top drawer", ``open the bottom drawer", ``close the bottom drawer" \\
\hline
\end{tabular}

\caption{Details of Simulator \& Real-world Manipulation Skills}
\label{tab:skill-details}
\end{table*}

\section{Dataset Details}\label{DD}

To validate the proposed LLCRM task, we construct a comprehensive lifelong robotic manipulation benchmark. The benchmark settings are shown in Figure 4 (main paper), including 12 common robotic manipulation skills based on the RLBench simulator \cite{RLBench, llarva} and 6 real-world manipulation skills. The first 16 skills serve as training and testing tasks for continual learning (simulator with 800 episodes of training for each skill, real-world with 100 episodes of training for each skill), while the last two skills are not trained and serve as open-world generalization testing. The agent is required to learn the 16 skill tasks in sequence, and then take a test on each skill after completing all the learning. Please note that in the LLCRM setting, the skill-id $t$ is agnostic during the testing phase. Agents need to autonomously load the knowledge they have learned in order to respond to current skill manipulations. We summarize the details of the proposed lifelong robotic manipulation benchmark in Table \ref{tab:skill-details}. Please note that to distinguish between real-world and simulation environment instructions, we follow LLARVA's prompt template \cite{llarva}: You are a Franka/UR5 robot using end-effector control. The task is ``xxx (user instruction)".

\section{Preliminary Study Implementation}\label{PS}

In this section, we provide details of our Preliminary Study experiment settings. To explore the skill-shared and skill-specific knowledge, we use LoRA \cite{lora}, i.e., a learnable $\Delta\textbf{W} = \textbf{A} \textbf{B}$ ($r=4$) to fine-tune the pretrained LLARVA \cite{llarva} with six robotic manipulation skills (Task1, Task3, Task5, Task7, Task9, and Task11 as summarized in Table \ref{tab:skill-details}) to obtain ten sets of LoRA $\{\textbf{A}_{1}\textbf{B}_{1}, \textbf{A}_{2}\textbf{B}_{2},...,\textbf{A}_{10}\textbf{B}_{10}\}$. 

For \textit{Observation 1}, we calculate ten skills' parameter subspace similarity matrices and semantic subspace similarity matrices. The parameter subspace similarity matrices measure the similarity $Sm(\cdot)$ between each of the skill parameter subspaces $\{\textbf{A}_{1}\textbf{B}_{1},\textbf{A}_{2}\textbf{B}_{2},...,\textbf{A}_{10}\textbf{B}_{10}\}$. The semantic subspace similarity matrices measure the similarity $Sm(\cdot)$ between each of the skill instruction semantic subspaces $\{E_{T}(\mathcal{I}_{1}), E_{T}(\mathcal{I}_{2}), ..., E_{T}(\mathcal{I}_{10})\}$, where $E_{T}$ is CLIP text encoder and $\mathcal{I}_{t}$ is user language instruction for $t$-th skill.

For \textit{Observation 2}, we calculate the average value of each skill parameter subspace in different transformer layers $\{\{\Delta\bar{\textbf{W}}_{1}\}^{L}_{l=1},\{\Delta\bar{\textbf{W}}_{2}\}^{L}_{l=1},...,\{\Delta\bar{\textbf{W}}_{10}\}^{L}_{l=1}\}$ to demonstrate the learning importance for skill-specific knowledge at every transformer level. 

For \textit{Observation 3}, we calculate the similarity $Sm(\cdot)$ between each $t$-th skill subspace $\textbf{A}_{t}$ and the other nine skills $\{\textbf{A}_{j}\}_{j \neq t}$, and the similarity $Sm(\cdot)$ between each $t$-th skill subspace $\textbf{B}_{t}$ and the other nine skills $\{\textbf{B}_{j}\}_{j \neq t}$, separately. Then, we summarize their average values to obtain average similarity for each skill $\{\bar{Sm}(\textbf{A}_{1},\{\textbf{A}_{j}\}),\{\bar{Sm}(\textbf{A}_{2},\{\textbf{A}_{j}\}),...,\{\bar{Sm}(\textbf{A}_{10},\{\textbf{A}_{j}\})\}$. 

Please note that the data in our Paper Figure 2 is normalized for better visualization.

\section{Ablation Studies Results and Analysis}\label{AS}

\begin{table*}[t]
\centering
\setlength{\tabcolsep}{1mm}
\renewcommand{\arraystretch}{1.3}
\caption{Ablation study for skills shared/special knowledge with ASR \%. }
\resizebox{\linewidth}{!}{
\begin{tabular}{l|ccc|cccccccccccccccccc|c}
\toprule
{Methods} & GGM & SOT & INA & S1 & S2 & S3 & S4 & S5 & S6 & S7 & S8 & S9 & S10 & S11 & S12 & S13 & S14  & S15 & S16 & S17 & S18 & ~Avg.~ \\
\midrule
Baseline & \xmarkg & \xmarkg & \xmarkg & 0 & 0 & 0 & 0 & 0 & 0 & 0 & 0 & 0 & 0 & 0 & 0 & 12 & 4 & 4 & 80 & 4 & 0 & 5.8 \\
SkillsCrafter w/o GGM & \xmarkg & \cmark & \cmark & 48 & 76 & 32 & 48 & 48 & 48 & 40 & 4 & 12 & 8 & 12 & 68 & 92 & 76 & 72 & 76 & 40 & 44 & 46.9 \\
SkillsCrafter w/o INA & \cmark & \cmark & \xmarkg & 52 & 72 & 28 & 52 & 52 & 48 & 44 & 12 & 12 & 12 & 8 & 72 & \textbf{84} & 80 & 80 & 84 & 32 & 44 & 48.2 \\
SkillsCrafter w/o SOT & \cmark & \xmarkg & \cmark & 56 & 80 & \textbf{40} & 60 & \textbf{52} & 44 & \textbf{48} & 8 & 16 & 8 & 8 & 68 & 84 & 80 & 72 & 76 & 36 & 48 & 49.1 \\
\midrule
\rowcolor{lightgray}
\textbf{SkillsCrafter} & \cmark & \cmark & \cmark & \textbf{56} & \textbf{80} & 36 & \textbf{64} & 48 & \textbf{52} & 44 & \textbf{12} & \textbf{20} & \textbf{12} & \textbf{12} & \textbf{72} & \textbf{92} & \textbf{84} & \textbf{80} & 80 & \textbf{40} & \textbf{52} & \textbf{52.0}  \\
\midrule
\end{tabular}}
\label{a1}
\end{table*}

\begin{table*}[t]
\centering
\setlength{\tabcolsep}{1mm}
\renewcommand{\arraystretch}{1.3}
\caption{Ablation study for skills shared/special knowledge with FR \%. }
\resizebox{\linewidth}{!}{
\begin{tabular}{l|ccc|cccccccccccccccccc|c}
\toprule
{Methods} & GGM & SOT & INA & S1 & S2 & S3 & S4 & S5 & S6 & S7 & S8 & S9 & S10 & S11 & S12 & S13 & S14  & S15 & S16 & S17 & S18 & ~Avg.~ \\
\midrule
Baseline & \xmarkg & \xmarkg & \xmarkg & 100 & 100 & 100 & 100 & 100 & 100 & 100 & 100 & 100 & 100 & 100 & 100 & 88 & 96 & 95 & 0 & 92 & 100 & 92.8 \\

SkillsCrafter w/o GGM & \xmarkg & \cmark & \cmark & 20 & 10 & 27 & 43 & 14 & 14 & 23 & 80 & 57 & 50 & 25 & 15 & 4 & 17 & 18 & 0 & 23 & 35 & 26.5 \\

SkillsCrafter w/o INA & \cmark & \cmark & \xmarkg & 13 & 14 & 36 & 38 & 7 & 14 & 15 & 40 & 57 & 25 & 50 & 0 & 13 & 13 & 9 & \textbf{0} & 38 & 35 & 23.9 \\

SkillsCrafter w/o SOT & \cmark & \xmarkg & \cmark & 7 & 5 & \textbf{9} & 29 & \textbf{7} & 21 & \textbf{8} & 60 & 43 & 50 & 50 & 15 & 13 & 13 & 18 & 5 & 31 & 29 & 22.9 \\
\midrule
\rowcolor{lightgray}
\textbf{SkillsCrafter} & \cmark & \cmark & \cmark & \textbf{7} & \textbf{5} & 18 & \textbf{24} & 14 & \textbf{7} & 15 & \textbf{40} & \textbf{29} & \textbf{25} & \textbf{25} & \textbf{10} & \textbf{4} & \textbf{9} & \textbf{9} & 0 & \textbf{23} & \textbf{24} & \textbf{16.0} \\
\midrule
\end{tabular}}
\label{a2}
\end{table*}

\begin{table*}[t!]
\centering
\setlength{\tabcolsep}{1mm}
\renewcommand{\arraystretch}{1.3}
\caption{Ablation study for skills Knowledge aggregation with ASR \%. }
\resizebox{\linewidth}{!}{
\begin{tabular}{l|cccc|cccccccccccccccccc|c}
\toprule
{Methods} & IM & VM & Avg & TOP & S1 & S2 & S3 & S4 & S5 & S6 & S7 & S8 & S9 & S10 & S11 & S12 & S13 & S14  & S15 & S16 & S17 & S18 & ~Avg.~ \\
\midrule
SkillsCrafter w/ TOP & \xmarkg & \xmarkg & \xmarkg & \cmark & 60 & 80 & \textbf{44} & \textbf{76} & \textbf{56} & 52 & \textbf{48} & \textbf{16} & \textbf{24} & 12 & \textbf{16} & 72 & 92 & \textbf{88} & 80 & 80 & 4 & 8 & 50.4 \\

SkillsCrafter w/ Avg & \xmarkg & \xmarkg & \cmark & \xmarkg & \textbf{60} & 80 & 40 & 64 & 44 & 48 & 44 & 12 & 16 & 8 & 16 & 68 & 88 & 40 & \textbf{84} & 72 & 8 & 8 & 44.4 \\

SkillsCrafter w VM & \xmarkg & \cmark & \xmarkg & \xmarkg & 52 & 76 & 32 & 60 & 44 & 52 & 48 & 16 & 24 & 12 & 8 & 76 & 84 & 88 & 84 & 76 & 36 & 12 & 48.9 \\

\midrule
\rowcolor{lightgray}
\textbf{SkillsCrafter} & \cmark & \xmarkg & \xmarkg & \xmarkg & 56 & \textbf{80} & 36 & 64 & 48 & \textbf{52} & 44 & 12 & 20 & \textbf{12} & 12 & \textbf{72} & \textbf{92} & 84 & 80 & \textbf{80} & \textbf{40} & \textbf{52} & \textbf{52.0} \\
\midrule
\end{tabular}}
\label{a3}
\end{table*}

\begin{table*}[t!]
\centering
\setlength{\tabcolsep}{1mm}
\renewcommand{\arraystretch}{1.3}
\caption{Ablation study for skills Knowledge aggregation with FR \%. }
\resizebox{\linewidth}{!}{
\begin{tabular}{l|cccc|cccccccccccccccccc|c}
\toprule
{Methods} & IM & VM & Avg & TOP & S1 & S2 & S3 & S4 & S5 & S6 & S7 & S8 & S9 & S10 & S11 & S12 & S13 & S14  & S15 & S16 & S17 & S18 & ~Avg.~ \\
\midrule
SkillsCrafter w/ TOP & \xmarkg & \xmarkg & \xmarkg & \cmark & 0 & 5 & \textbf{0} & \textbf{10} & \textbf{0} & 7 & \textbf{8} & \textbf{20} & \textbf{14} & 25 & \textbf{0} & 10 & 4 & \textbf{4} & 9 & 0 & 92 & 88 & 16.5 \\
SkillsCrafter w/ Avg & \xmarkg & \xmarkg & \cmark & \xmarkg & 0 & 5 & 9 & 24 & 21 & 14 & 15 & 40 & 43 & 50 & 0 & 15 & 8 & 57 & \textbf{5} & -5 & 85 &88 & 26.3 \\
SkillsCrafter w VM & \xmarkg & \cmark & \xmarkg & \xmarkg & 13 & 10 & 27 & 29 & 21 & 7 & 8 & 20 & 14 & 25 & 50 & \textbf{5} & 13 & 4 & \textbf{5} & 0 & 31 & 82 & 20.2 \\
\midrule
\rowcolor{lightgray}
\textbf{SkillsCrafter} & \cmark & \xmarkg & \xmarkg & \xmarkg & \textbf{7} & \textbf{5} & 18 & 24 & 14 & \textbf{7} & 15 & 40 & 29 & \textbf{25} & 25 & 10 & \textbf{4} & 9 & 9 & \textbf{0} & \textbf{23} & \textbf{24} & \textbf{16.0} \\
\midrule
\end{tabular}}
\label{a4}
\end{table*}

\textbf{How Do Skills Shared and Special Knowledge Perform?} The detailed ablation studies results about skills shared and special knowledge are summarized in Table \ref{a1} and Table \ref{a2}. The “GGM” denotes the Gumbel-softmax Gating Mechanism based sparse loading strategy, which enhances the exploration of the skill-special knowledge; The “INA” denotes the Inheritance $\textbf{A}$ strategy, which enhances the exploration of the skill-shared knowledge; The “SOT” denotes the Skills Orthogonal Term, which also enhances the exploration of the skill-special knowledge and consolidates the structure of LoRA's skill-shared and skill-special knowledge. The ablation results demonstrate the effectiveness of each part of SkillsCrafter, effectively reducing catastrophic forgetting (Table \ref{a1}) and improving new skill learning (Table \ref{a2}). \textbf{Analysis 1:} GGM assigns different learning weights to each manipulation skill in different layers through self-learning. Compared to loading learning weights in all layers, this dynamic sparse approach is more flexible and helps learn skill-specific knowledge, thereby achieving better performance. \textbf{Analysis 2:} INA provides prior knowledge for learning each skill. Specifically, it inherits the skill knowledge most similar to the current skill from previous learning skills to provide a better initial value for learning new skills. Therefore, this method of exploring shared knowledge between skills achieves better performance. \textbf{Analysis 3:} SOT assigns specific orthogonal subspaces to each skill, avoiding interference between skills and thus achieving better results.

\textbf{How Do Skills Knowledge Aggregation Perform?} The ablation studies results about the knowledge aggregation methods are summarized in Table \ref{a3} and Table \ref{a4}. The “IM” denotes the Knowledge aggregation based on Instruction Matching; The “VM” denotes the Knowledge aggregation based on Vision Matching; The “Avg” denotes the average pooling for all instruction encodings, rather than using SVD; The “TOP” denotes the selection of the most similar LoRA in SkSA. Based on the ablation results, we provide three insightful analyses. \textbf{Analysis 1:} Compared to the VM method, the IM method we used achieves more effective results. We think this is because user instructions contain all the semantic information about the manipulative skill, while vision only contains information about the appearance of the object and cannot provide specific information for a specific robotic task, making it difficult to accurately aggregate previously learned knowledge. \textbf{Analysis 2:} Compared to the Avg method, we achieve better performance using the SVD method. We think this is because SVD extracts commonalities from diverse instructions and describes the essential semantic space of skills, rather than simply taking the average of diverse instructions. \textbf{Analysis 3:} Compared to the TOP method, we achieve better results in open-world tasks (S17, S18) using the aggregation method. We think this is because the aggregation method soft-aggregates all the important knowledge learned previously when faced with unknown tasks, rather than choosing the single most relevant knowledge in an absolute way, thus achieving better generalization. For example, S17 involves stacking wooden blocks in a real-world environment. If the TOP method is used, it can only access the most similar knowledge from S11. However, S17 uses the real-world environment of UR5, while S11 uses the simulated environment of Franka, and the differences between the two are significant. Therefore, it is necessary to access the diverse knowledge learned previously to achieve optimal generalization performance.

\section{Manipulation Platform}\label{UR}

\begin{figure}[!t]
\centering
\includegraphics[width=0.85\linewidth]{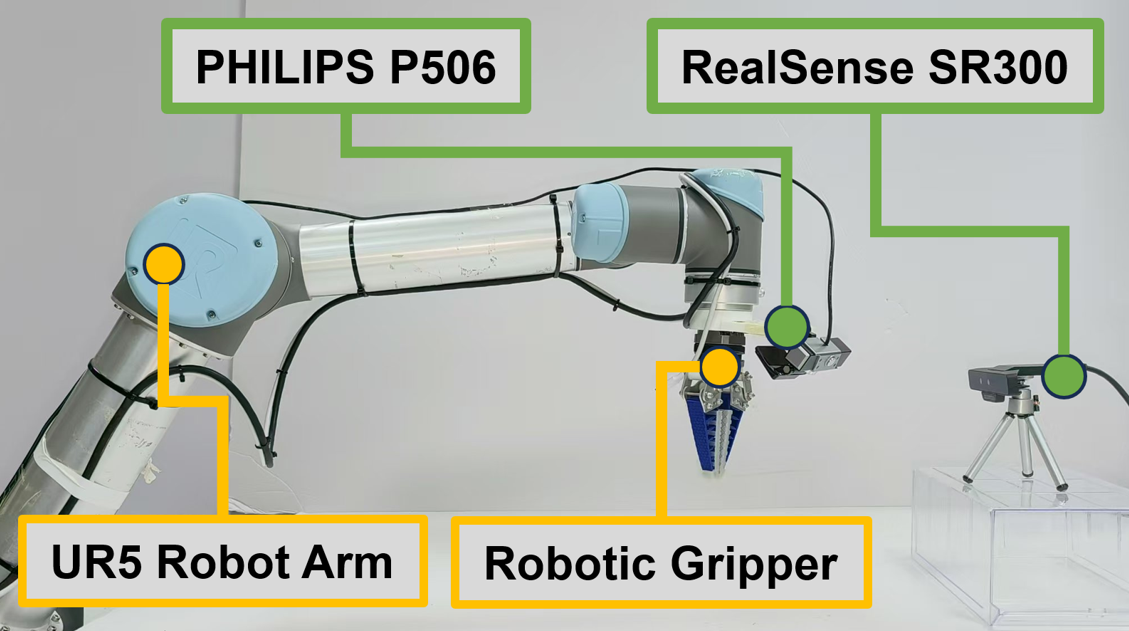}
\caption{A photograph of the proposed UR5 robotic system. 
}
\label{fig7}
\end{figure}

\begin{figure}[!t]
\centering
\includegraphics[width=1\linewidth]{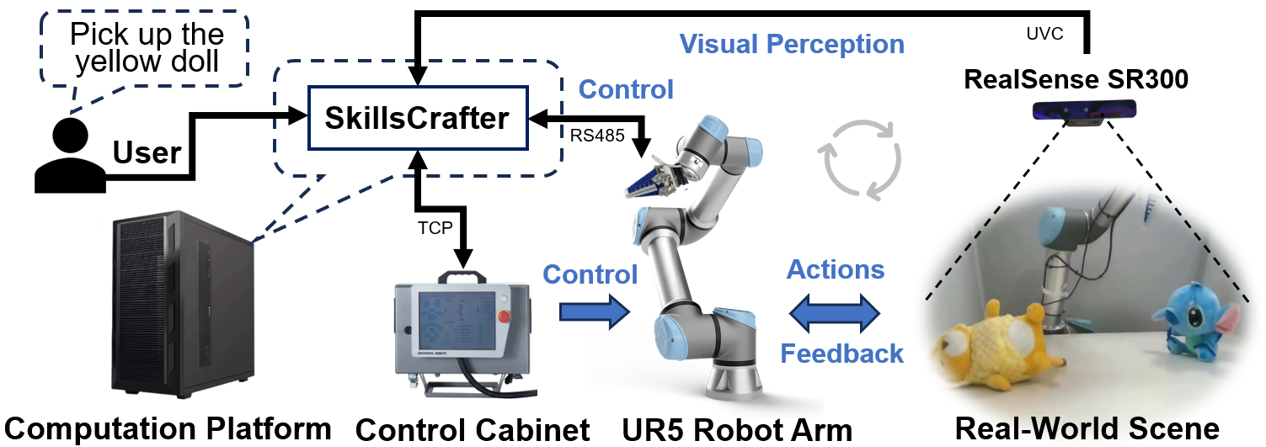}
\caption{Illustration of the proposed UR5 robotic system. 
}
\label{fig8}
\end{figure}

The proposed UR5 robotic system is shown in Figure \ref{fig7}. The system consists of a UR5 robotic arm, a RealSense SR300 camera, a PHLIPS P506 camera, a control cabinet, and a calculation platform, as shown in Figure \ref{fig8}. The proposed system enables users to issue natural language instructions, such as “Pick up the yellow doll,” which are processed by a central module named SkillsCrafter based on a computation platform. SkillsCrafter interprets user intents and coordinates control signals between the computational platform, the robot control cabinet, and the UR5 arm via TCP and RS485 protocols. A RealSense SR300 camera captures visual input from the real-world environment, providing critical perception data that enables scene understanding. The visual data is sent to the SkillsCrafter module using the UVC protocol. Based on this perception, SkillsCrafter generates appropriate actions that are executed by the UR5 robot arm, which interacts with the environment and provides feedback for continuous refinement. This closed-loop system tightly integrates perception, control, and reasoning to support complex, user-driven manipulation tasks.

\end{document}